\documentclass[10pt,twocolumn,letterpaper]{article}

\usepackage[utf8]{inputenc}
\let\labelindent\relax
\usepackage{nccv}
\usepackage{times}
\usepackage{epsfig}
\usepackage{graphicx}
\usepackage{amsmath}
\usepackage{amssymb}
\usepackage{datetime}
\usepackage{enumerate}
\usepackage[official]{eurosym}
\usepackage{float}                
\usepackage{xtab}                 
\usepackage{verbatim}
\usepackage{framed}
\usepackage{listings}        
\usepackage{footnote}             
\usepackage{array}                
\usepackage{xspace}               
\usepackage{siunitx}              
\usepackage[english]{babel}
\usepackage{csquotes}
\usepackage{framed}
\usepackage{xcolor}
\usepackage{tabto}
\usepackage{pdfpages}
\usepackage{textcomp,gensymb} 
\usepackage[pagebackref=true,breaklinks=true,letterpaper=true,colorlinks,bookmarks=false]{hyperref}
\usepackage[nameinlink]{cleveref}
\usepackage{regexpatch} 
\usepackage{etoolbox}
\usepackage{float}
\usepackage[caption = false]{subfig}
\usepackage{mathtools}
\usepackage[numbers]{natbib}
\DeclarePairedDelimiter\abs{\lvert}{\rvert}%
\DeclarePairedDelimiter\norm{\lVert}{\rVert}%

\makeatletter
\let\oldabs\abs
\def\abs{\@ifstar{\oldabs}{\oldabs*}}
\let\oldnorm\norm
\def\norm{\@ifstar{\oldnorm}{\oldnorm*}}
\makeatother

\makeatletter

\makeatletter
\xpatchcmd{\@todo}{\setkeys{todonotes}{#1}}{\setkeys{todonotes}{inline,#1}}{}{}
\makeatother

\makeatletter
\renewcommand{\fps@figure}{htbp}
\renewcommand{\fps@table}{htb}
\makeatother

\graphicspath{ {./figures/} }

\usepackage{blindtext}
\usepackage{graphicx}
\usepackage[justification=centering]{caption}
\usepackage{url}

\usepackage{amsmath} 

\usepackage{booktabs} 

\newcommand{\pe}{Proteus-Eretes }
\nccvfinalcopy 

\def\nccvPaperID{39} 

\ifnccvfinal\fi
\begin{document}

\title{ Fine-grained Classification of Rowing teams}

\author{L.J. Hamburger\\
Delft University of Technology\\
2600 GA Delft, The Netherlands\\
{\tt\small l.hamburger@student.tudelft.nl}\\
\and
M.J.A. van Wezel\\
Delft University of Technology\\
2600 GA Delft, The Netherlands\\
{\tt\small m.j.a.vanwezel@student.tudelft.nl} \\ 
\and
Y. Napolean\\
Delft University of Technology\\
2600 GA Delft, The Netherlands\\
{\tt\small Y.Napolean@tudelft.nl}
}

\markboth{TU DELFT - CS4180 PAPER}%
{Shell \MakeLowercase{\textit{et al.}}: Bare Demo of IEEEtran.cls for Journals}

\maketitle
\thispagestyle{plain}
\pagestyle{plain}
\begin{abstract} 
\label{sec:abstract}

Fine-grained classification tasks such as identifying different breeds of dog are quite challenging as visual differences between categories is quite small and can be easily overwhelmed by external factors such as object pose, lighting, etc. This work focuses on the specific case of classifying rowing teams from various associations.

Currently, the photos are taken at rowing competitions and are manually classified by a small set of members, in what is a painstaking process. To alleviate this, Deep learning models can be utilised as a faster method to classify the images. 
Recent studies show that localising the manually defined parts, and modelling based on these parts, improves on vanilla convolution models, so this work also investigates the detection of clothing attributes.

The networks were trained and tested on a partially labelled data set mainly consisting of rowers from multiple associations.

This paper resulted in the  classification of up to ten rowing associations by using deep learning networks the smaller VGG network achieved 90.1\% accuracy whereas ResNet was limited to 87.20\%. Adding attention to the ResNet resulted into a drop of performance as only 78.10\% was achieved.
\end{abstract}



\section{Introduction}\label{sec:Introduction}

At every rowing competition, each rowing association takes many pictures to document the event. However, this results in many hours of work and is also an error-prone process.

Therefore it would be interesting to see how well deep-neural networks behave when faced with classes containing very few distinguishing features.


In these images, each rowing team has certain visual features that allow one to determine their association: there are the uniforms, the colour of the blades, and the boats themselves. As clothing is the most distinct of these attributes, it was logical to focus on networks that have been optimised for clothing. 
The same boat can be used by other association throughout an event. 
Therefore determine the association by the boat would be error-prone.

\begin{figure}[th!]
  \includegraphics[width=0.49\linewidth,keepaspectratio]{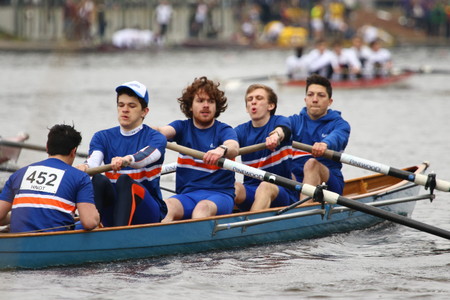}
    \includegraphics[width=0.49\linewidth,keepaspectratio]{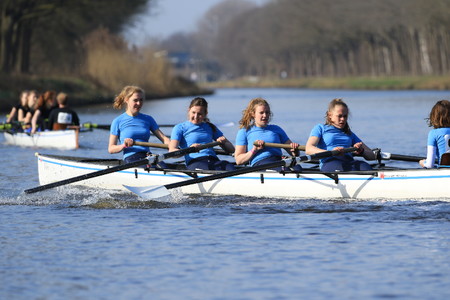}
    \caption{An example image from our rowing boat data set. Here a crew from \pe and Njord is visible. Under specific lighting conditions, the different shades of blue can appear similar. The similarities will make it difficult for a deep learning network to correctly classify the teams.
    }
  \label{fig:Heatmap of proteus}
\end{figure}


As  images between classes can appear similar, the research question becomes: How do classifiers behave when faced classes with few distinguishing features?

In this paper is a new type of dataset is described and there is looked into how classifiers behave when faced classes with few distinguishing features.

Several different network architectures were investigated such as VGGNet, ResNet and Inception-ResNet which will be further discussed in \cref{sec:Network}.

As attention is used in several networks which improve the accuracy of the trained network \cite{Ba2014MultipleOR, DBLP:journals/corr/XuBKCCSZB15}. In \cite{WangJQYLZWT17} several limitations are noted which can  impact  the performance with our data set. For instance, the images of the boats are taken from several angles and contain varying shapes for each class.

To the best of the author's knowledge, there has not been much work regarding how attention models work in the context of fine grained classification, that is the focus of this body of work.

Our main contributions are: 

\begin{itemize}
    \item Comparison of the performance in fine grained classification of rowing teams for several network architectures.
    
    \item Investigating the performance change that incurs in the aforementioned models when subject to various inputs (changing image size, no. of images).
    
    \item Analyse the performance of attention based models in the context of fine grained classification.
\end{itemize}

This report first discusses the existing body of work that is related to this research in \cref{sec:relatedwork} and also presents relevant existing data sets, while \cref{sec:Dataset} describes how the \pe data set looks like. Additionally, possible improvements to this data set will be discussed. The section \Cref{sec:Network} focuses on the different networks considered and compared, as well as how their performance is, on our data set. Then, \cref{sec:Results} presents the results of experiments evaluating the networks. Finally the report is concluded in \cref{sec:Conclusion}, and further work is discussed in \cref{sec:Discussion}.

\section{Related work}\label{sec:relatedwork}

Considering the fact that teams in the rowing competition usually have different clothing, detecting these clothing attributes could be one possible solution to classifying these images. Research on clothing attribute detection typically utilises datasets such as Deep Fashion \cite{liuLQWTcvpr16DeepFashion} and Fashion-MNIST. Deep fashion and Fashion-MINST are databases of clothes and state of the art models performance on these show that it is possible that a deep learning algorithm is able to learn features of different types of clothing and differentiate them from each other \cite{liuLQWTcvpr16DeepFashion, zalando_2018}. However, given that clothing can appear similar across classes, rather than focusing specifically on clothing, this research casts this as a fine grained classification problem.\newline

There has been quite some research on fine-grained classification. There are several data sets available for fine-grained classification such as stanford-cars \cite{krause20133d}, and NAbirds \cite{van2015building}. Several models such as FixSENet-154 \cite{DBLP:journals/corr/abs-1906-06423} and WS-DAN \cite{hu2019see} have remarkable performance in the aforementioned data sets. However, to the best of the authors knowledge not much work has been done to understand how classifiers behave in the context of fine grained classification. 

When considering coarser classification, ResNet \cite{he2016deep} is the one of the best performing networks and VGGNet \cite{simonyan2014very} has also shown to learn features significant for tasks such as style transfer \cite{gatys2016image}
A more in-depth explanation of the networks investigated can be found in \cref{sec:Network}. \newline

In the paper, attention is all you need \cite{NIPS2017_7181}, it has been shown that the attention module improves  performance of a vanilla CNN. The attention module helps the deep network focus on semantically relevant parts and remove the uninteresting clutter of the image \cite{Han:2018:AAM:3240508.3240550}.
There are not many studies that look into how attention models work in fine-grained classification. In some work \cite{Xiao_2015_CVPR} the focus is on the architecture instead of analysing how the network improved in performance.  \newline 

Different type of papers that are about rowing boats or ships are mostly about detecting the ship or are about having sensors placed that detect movements of the persons rowing \cite{twentenRoeiBootje} or detecting ships in a shipyard \cite{DBLP:journals/corr/abs-1806-04331}. 

To the best of the authors knowledge, there is no study found that is based on detecting rowing teams. It was expected that big events like the Tour de France has methods published about detecting  different type of associations. In this kind of events they mostly use transceivers \cite{toerdebier} to determine which member is driving where.


\section{Data set} \label{sec:Dataset}
For this project the organisation \pe supplied the authors a total of nearly 67 thousand images, from which the data set is constructed. 
The given images are scaled down to 450 x 350 pixels with a total of 71 rowing associations (teams). 
In \cref{tab:data_set} the division of the data set is shown for the top 21 associations in terms of images that are labelled. 
It is clear that the size of the data set is unevenly distributed and contains a large amount of unlabelled data.
In order to ensure that the network trains correctly and does not get biased towards any one association, an equal distribution of photos per association (classes) is used. 
This choice led to ten usable classes with $11.480$ images available for training.
Appendix \ref{FirstAppendix} gives a general impression of the data set, this by displaying several exemplary images of the different classes. 

\begin{figure}[th!]
  \includegraphics[width=\linewidth,keepaspectratio]{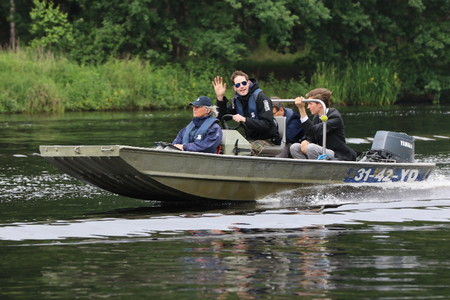}
  \caption{Image which is wrongly labelled as Proteus in the data set. Which could result a performance drop of the final network.}
  \label{fig:random_image1}
\end{figure}

Since the authors are the first users of the data set, it is worth mentioning the associated challenges. 
First of all, some rowing associations also have unique clothing attributes for each fifth anniversary. 
In this case, the rowing association may change their club colours leading to a more challenging classification due to multiple unique identifications per class. 
Not every association uses this possibility, so there are classes with a single or multiple unique identifications.
Another challenging characteristic is the presence of images that capture multiple rowing associations. 
Multiple teams in one image could be problematic since only one label is known, increasing the risk of possible miss-classification.
Next to the challenges, also part of the data set is incorrectly labelled as shown in \cref{fig:random_image1}.
Extensive research on the data set characteristics has shown that about 5\% of each class is incorrectly labelled.

It is also possible that the data consists of images that no team is present in the image. It is expected that this type of randomly labelled images will hurt the overall performance, this due to the increased Bayes error. However, as shown in \cref{sec:Results}, an accuracy of 90\% can be achieved. 

\begin{table}
    \centering
    \caption{Overview of the distribution of the classes within the data set. It is clear that the original data set is highly unbalanced and that a large amount of images do not contain any labels.}
    \label{tab:data_set}
    \begin{tabular}{lrlr}
        \toprule
        \textbf{Label} &  \textbf{\#} & \textbf{Label} &  \textbf{\#} \\ 
        \midrule
        Unknown & 41522 & Phocas & 943 \\ 
        \pe & 2667 & Aegir & 924 \\ 
        Gyas & 2112 & Theta & 906 \\ 
        Skoll & 1974 & Okeanos & 846 \\ 
        Orca & 1559 & Asopos de Vliet & 840 \\ 
        Laga & 1543 & Euros & 819 \\ 
        Nereus & 1506 & Vidar & 769 \\ 
        Triton & 1440 & Saurus & 582 \\ 
        Njord & 1354 & Pelargos & 423 \\ 
        Skadi & 1172 & Boreas & 211 \\ 
        Argo & 1148 & De Maas & 170 \\ 
        \bottomrule
    \end{tabular}
\end{table}

\subsection{Test set rowing images}
In order to evaluate the performance of the network a test set withheld during training is included in the data set. The test data is independent of the training data, but that follows the same probability distribution as the training data. This step is crucial to test the generalizability of the model. For the associations Skadi and Argo new images were carefully gathered form the unlabelled data set, as they span the various classes that the model would face, when used in the real world. For the other classes label data was already available. The size of the test set is 1000 images 100 for each association.

\section{Network}\label{sec:Network}

Due to the nature of the chosen application, detecting rowing associations from the given images, the use of Convolutional Neural Network (CNN, or ConvNet) is a natural fit. A CNN is a special kind of multi-layer neural network, designed to recognise visual patterns directly from pixel images with minimal preprocessing. 

The most obvious way to recognise the rowing associations is by the clothing attributes of the rowers. 
Previous research, has shown that a more general direction is beneficial. 
Based on the results of the ImageNet competition, a selection of suitable networks is chosen.

The first candidate is the ResNet architecture since it achieves a high performance in classification tasks. 
For the second architecture, the VGGNet network is selected. 
Moreover, there was an additional test done with Inception ResNet which has previously showed promising results.
 
The network architectures can be easily modified for this application which is beneficial since the ImageNet data set is significantly more complex in comparison to this application, they have 1000 categories as compared to only 10 classes that we use.
Due to the reduced complexity of this application, down-scaling the architecture could be worthwhile in order to reduce the possibility of overfitting and save training time. 
As another option that is looked at is Inception-ResNet-12 which is a version of ResNet. This is also one of the top competitors as shown in \cite{SzegedyIV16}. 
Following this, attention is also of interest as it has been shown to improve the performance of the previously discussed networks. 

\begin{figure}[th!]
  \includegraphics[width=\linewidth]{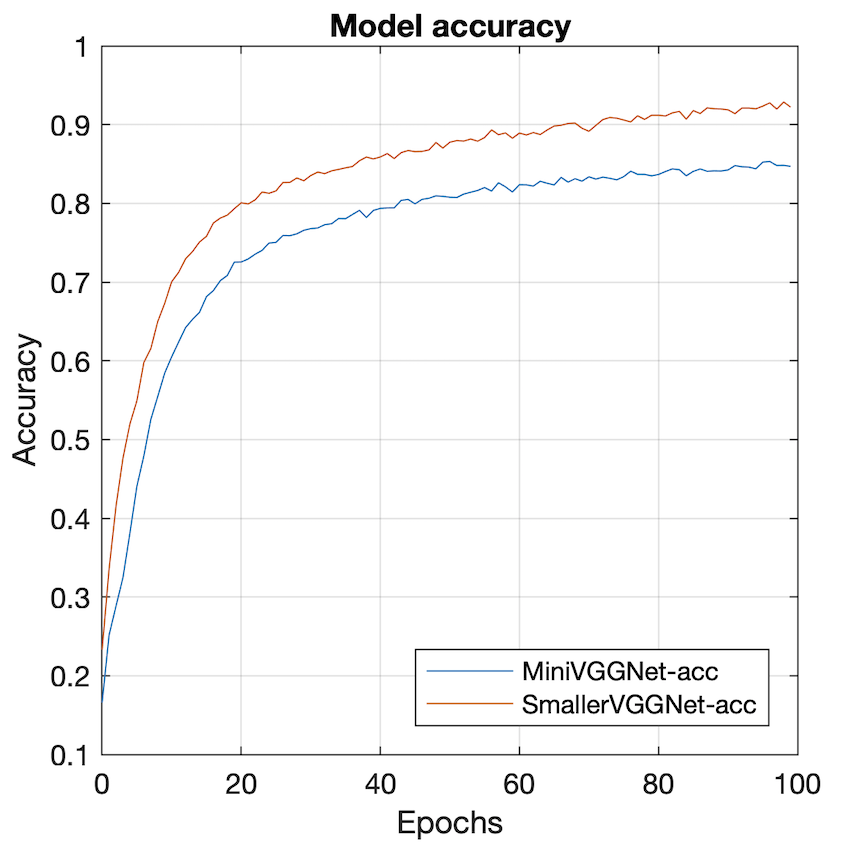}
  \caption{Comparing the performance difference of MiniVGG and SmallerVGG network. The SmallerVGG network clearly outperforms MiniVGG.}
  \label{fig:MiniVGG_vs_SmallerVGG}
\end{figure}

\subsection{VGGNet} \label{section:VGGNet}

In addition to performing well in style transfer tasks VGGNet does not have any skip connections like ResNet and is compartively smaller in size. This aids in the endeavour of comparing various types of architecture in the fine-grained classification setting
The simple structure is important since the ImageNet contest is a significantly more complex challenge in comparison to this application; for example, the ImageNet is based around 1000 categories compared to only ten categories. 
Due to this dramatic difference in complexity, a simplified network might be worthwhile to investigate. This simplification can be easily achieved due to the uniform structure of VGG.

The research of network similar to VGGNet has led to three possible candidates namely, VGGNet(16), MiniVGGNet \cite{MiniVGGNet}, SmallerVGGNet \cite{SmallerVGGNet}. VGGNet(16) is the full VGG network of 16 convolutional layers used in the ImageNet competition, SmallerVGGNet is a reduced version and has a depth of five convolutional layers, and MiniVGGNet is the smallest version with only four convolutional layers. Due to the increased  training time of larger models like the ResNet architecture, the choice was made to start small and expand when necessary. 

Testing these networks has shown that the MiniVGGNet and SmallerVGGNet already reach similar performance to the ResNet architecture (accuracy in the range of 70\% to 85\% on runs of 100 epochs), omitting the need for the full VVGNet(16). Furthermore, it achieved its performance in reasonable training time of less than 5 hours. The performance differences between the other two architecture are shown in \cref{fig:MiniVGG_vs_SmallerVGG}. The results show that SmallerVGGNet is the preferred VGG network due to its balance in performance and training time.

\subsection{ResNet} \label{section:ResNet}

ResNet is a shorthand version of Residual network.
The improved version of ResNet V2 was used when training.
ResNet has several different depths \cite{DBLP:journals/corr/HeZRS15, DBLP:journals/corr/HeZR016} where the smallest is 11 and size isn't limited but it will require more data to be trained and will have more issues with overfitting when using small data sets. 
The amount of data limits the performance of ResNet-56 as the accuracy was inconsistent, which is also seen in \cref{table:networkComparesson}.

\subsection{Inception ResNet} \label{section:InceptionResNet}

As shown in the paper \cite{SzegedyIV16} the Inception ResNet version outperforms ResNet and the network which only uses Inception.
Inception ResNet-V2 has increased the model size and lower computational efficiency and low parameter count are enabling better performance when having fully-connected layers \cite{DBLP:journals/corr/SzegedyVISW15}. 
The promising result from the field was that this network was considered.
It was visible that the chosen model required more images than the data sets could provide.
Due training the accuracy graphs of the network showed performance fluctuation while training.

\subsection{Attention} \label{subsection:NETattention}

Attention was introduced to help find the parts of the input where the deep neural network should focus on, the parts that have more semantic value.  
An attention block has an encoder-decoder structure, as shown in paper \cite{WangJQYLZWT17}. The paper also implemented an attention layer that is used for the implementation.
Multiple configurations of ResNet were already tested and ResNet-11 showed the best result. Therefore the ResNet-11 was used with the attention layer. \newline

The attention layer focuses the network on the rowing boat allowing the ResNet to improve its focus, and thus reducing background noise. The disadvantage is that it is resource hungry and can be  more time-consuming. Within our data set, the images are taken in different angles making it hard to for the model to focus on the boat itself. The results of attention on our networks will be further discussed in \cref{subsection:RESattention}.

\section{Results}\label{sec:Results}
To evaluate the performance of the networks on the data set several experiments have been run. Several different parameters have been tested, such as the image size to be used to the required amount of images and then comparing the networks.

\subsection{Image sizes}
\begin{figure}[hbt!]
  \includegraphics[width=\linewidth]{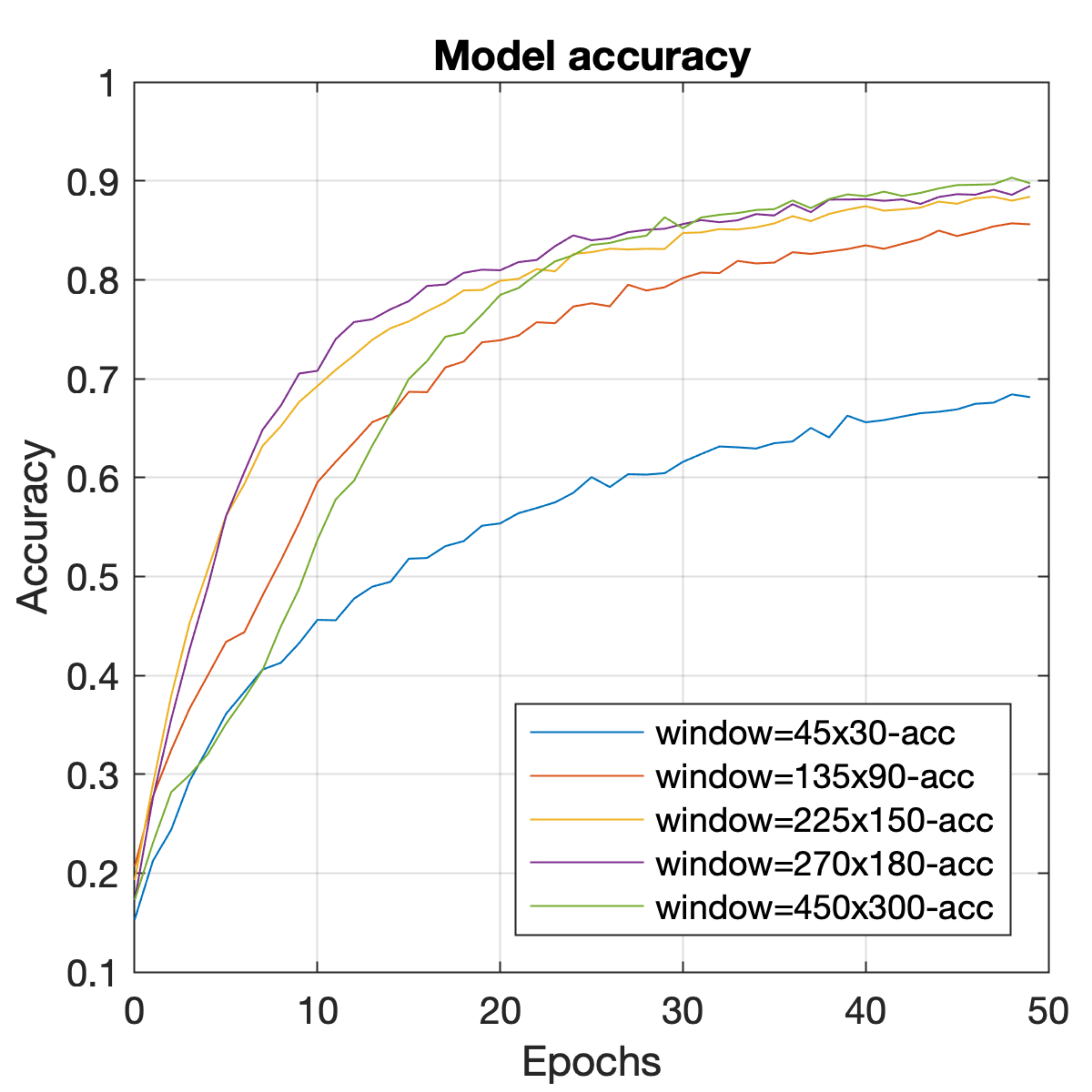}
  \caption{Training the network with different images size and the resulting performance. The sweet post is around a down scaled image to 225 by 150 instead of the original 450 by 350 pixels}
  \label{fig:Window_size}
\end{figure}

As the images contain a certain amount of information it is necessary to find the ideal images sizes. Too small an image would make learning impossible and too large will allow for over-fitting and large performance penalties. It was found that a smaller input size leads to lower performance, the most significant difference is nearly 20\% in comparison with larger image sizes. The impact of the input size is depicted in \cref{fig:Window_size}. 

The results show that a larger input size will lead to higher accuracy. A possible explanation is that more information and detail is extracted by the network leading a more reliable classification. Furthermore, it shows that there is no significant performance difference between a scaled input size of 270 by 180 pixels and the full resolution of 450 by 300 pixels. The benefit of the scaled input size is its strongly reduced training time and reducing the required amount RAM on the GPU.

\subsection{Required amount of images}
\begin{figure}[hbt!]
  \includegraphics[width=\linewidth]{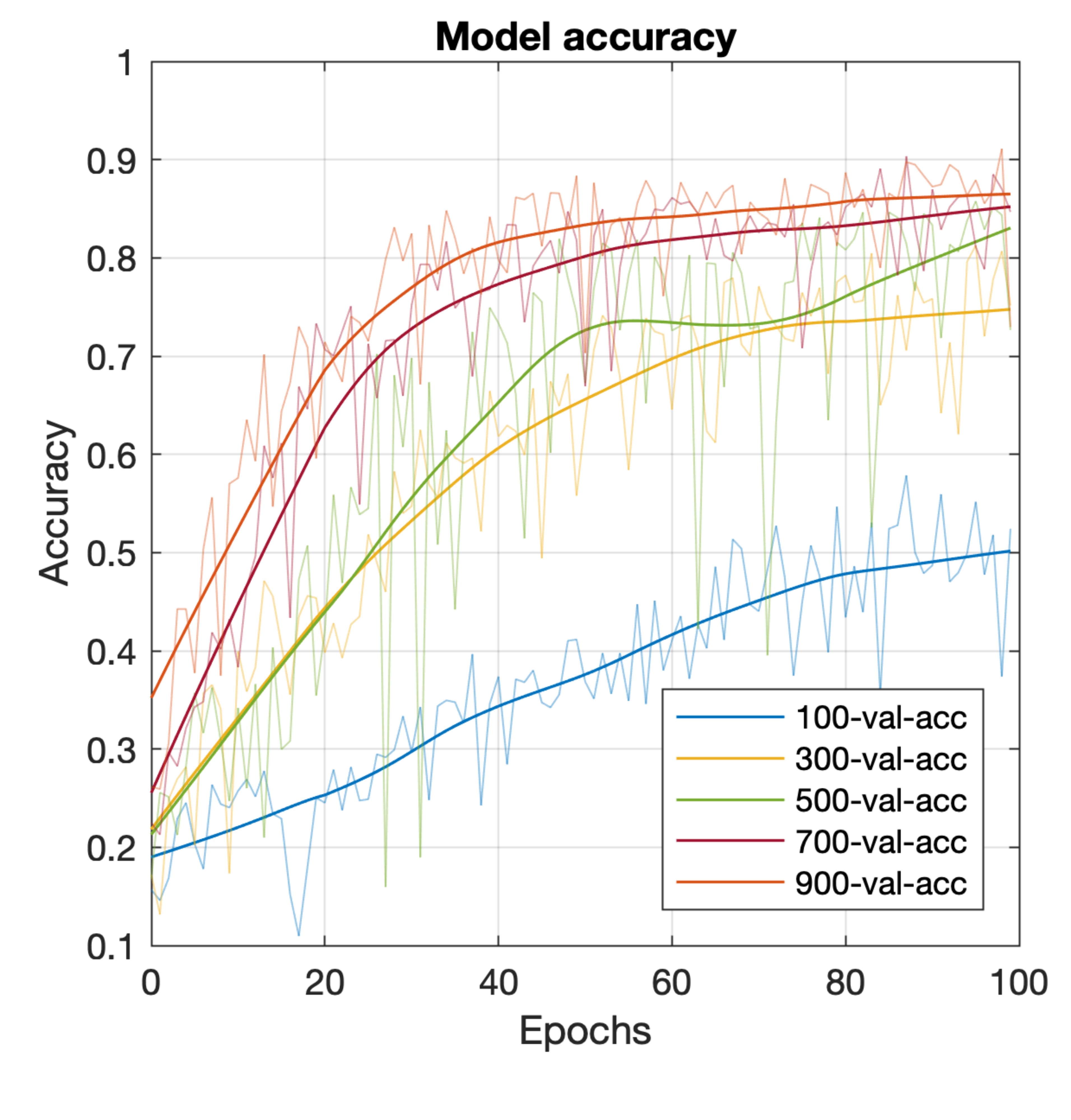}
  \caption{Training the network with different number of images in the data set. Verify that the data set is large enough and 700 images no large performance improvement is shown.}
  \label{fig:Dataset_size_performance}
\end{figure}

To verify if enough data is available and the learning works stable different amounts of input data  have been used and the performance has been assessed. In \cref{fig:Dataset_size_performance} the different of data set size are shown. 

The results depicted in \cref{fig:Dataset_size_performance} shows the effects of the different training sizes. The validation loss in particular shows that the smaller training set size, in the range of 100-500 image, suffers from overfitting. When more images are used the network no longer suffer from the same overfitting which is shown with the smaller data size. Higher performance might be achievable with a larger data set as the results clearly show that a larger data set size leads to higher achieved accuracy.

\subsection{Heatmap}\label{subsec:Heatmap}
To evaluate if a deep neural network learns on clothing attributes of rowing associations can be done using heatmaps.
Heatmaps were made using Lime \cite{marcotcr_2019, lime}. The heatmap shows how a particular network is detecting certain features of the data set without having to change the network. 
Considering for Class activation maps (CAM) \cite{gill_2016} it was found out the network had to be changed in a certain way to visualise how the networks classify the images. For that reason this was not used but Lime was used instead.\newline
In \cref{fig:Heatmap of proteus} the original images was showed and in \cref{fig:Heatmap of proteus attention} heatmaps without and with attention are showed.
From this in \cref{fig:Heatmap of proteus attention} shows how a network without attention that had a prediction certainty of 92\% react on this image.
From this image it is visible that our deep neural network learns on clothing attributes as was the primary purpose of the networks. 
While this image shows that another rowing association is in the same image the network focuses on the association in the front. This is indeed what was wanted as a single image was only supposed to be classified to a single class in the current network.

\subsection{Attention}\label{subsection:RESattention}
\begin{figure}[th!]
  \includegraphics[width=\linewidth,height=5.5cm]{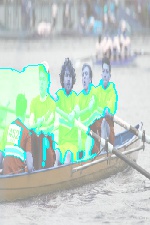}
 \includegraphics[width=\linewidth,height=5.5cm]{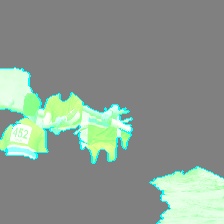}
  \caption{Heatmap created using the ResNet network. The first is without the attention module and the second one is with the attention module. Attention is reducing is using a smaller part of the image and has a bit more trouble focusing on the rowers.
  The original image is shown in \cref{fig:Heatmap of proteus}}
  \label{fig:Heatmap of proteus attention}
\end{figure}

The paper \cite{WangJQYLZWT17} already showed that occlusion is still a problem. 
The rowing boat data set has many occlusions and different camera poses, that could be the issue of why the performance diminishes. 
As showed in \cref{subsec:Heatmap} the deep neural network already learned a focus on the rowers. 
This layer of attention would be feasible if the images of rowers are consistent.
A heatmap is also created for the network with attention to look at how the network performed. In \cref{fig:Heatmap of proteus attention} it is shown that the attention layer actuality degrades the performance of the network when comparing with the original heatmap.

To verify if attention works correctly two experiments where run with the attention network on a subset of CIFAR-100. This conceited of the class trees and vehicles. Each of the set then consists of 5 classes with 2500 training images and 500 validation images are available. The trees where chosen due to the fact that the images are very much alike as what is the case with the rowing images. The other class vehicles have very clear distinct features for the network to learn and separate.

\begin{table}[ht!]
    \caption{The attention module tested with tree class from cifar-100. A CIFAR-10 data set with attention modules enabled from the paper \cite{WangJQYLZWT17} and in our implementation both at epoche 60.}
    \begin{tabular}{lcc} \toprule
    \textbf{Network}  & \textbf{Acc [\%]} \\ \midrule
    Trees                       & 54.8         \\
    CIFAR-10 paper              & 89.6         \\
    CIFAR-10 own                & 89.3       \\ \bottomrule
    \end{tabular}
    \label{table:networkComparesson:Attention}
\end{table}

When training ResNet without attention on the trees the network performed slightly better than when enabling the attention module. For the network with attention, as showed in table \cref{table:networkComparesson} the accuracy was 54.8 \% and without the accuracy of 56.2 \% was achieved . The tree class of CIFAR-100 consist of images with few distinguishing features.

To verify the attention network our implementation was compared with the original paper as shown in \cref{table:networkComparesson:Attention}.

\begin{table}[ht!]
\centering
\caption{Overview of top-performing networks architectures. At current smaller VGG is the best where as larger networks have more problems with overfitting}
\begin{tabular}{lcc} \toprule
\textbf{Network}  & \textbf{Val acc [\%]} & \textbf{Test acc [\%]}\\ \midrule
RESNET-11              & 93.6          &  87.2 \\
RESNET-11 \small{feature wise} & 86.2  &  81.6 \\
RESNET-11 attention    & 89.7          &  78.1 \\
RESNET-18              & 91.8          &  85.8 \\
RESNET-56              & 91.0          &  79.7 \\
Inception-RESNET-V2    & 92.1          &  76.1 \\
\textbf{SmallerVGG}    & \textbf{95.5} &  \textbf{90.1} \\
MiniVGG                & 87.3          &  81.9\\ \bottomrule
\end{tabular}
\label{table:networkComparesson}
\end{table}


\subsection{Network performance}

Multiple network architectures are trained on the described data set the results are depicted in \cref{table:networkComparesson}. The results show that the networks can deliver reliable performance, given the challenges to which the data set is subjected. 
The small performance differences between the test and validation data show that the network generalises well on unseen data. Different versions of the ResNet architecture (plain ResNet, ResNet with feature-wise normalized data augmentation, ResNet with attention) were investigated. The plain ResNet outperforms the modified ones. It is possible that the modifications cause the model to overfit.
The best performing network is the SmallerVGG architecture, this result was unexpected since earlier research, \cref{sec:relatedwork}, had shown that ResNet is the most favourable. When adding the attention module to the ResNet network a performance drop is seen. Attention ideally helps find the parts of the input which are more relevant for the network which reduces the required training time. Attention improves the performance on the validation data, but it shows a huge performance drop when run on the test set.

\section{Recommendations for Future Work} \label{sec:Discussion}

Training on more than ten different associations wasn't done because the data set is constrained by  the size of the smaller classes within the data set.
Training on more classes will require more data and cleaning of the current data by removing the wrongly labelled images and the removing of the lustrum year of the associations.
Using data augmentation can allow the creation of new images which can be used to train the network.
Data augmentation would consist of changing the lighting, cropping and perspective transform. 
The current implementation doesn't support multi-label classification the current data set doesn't contain multiple labels.
In several images, multiple associations are pictured, the network should be able to classify the different teams in the picture.
As attention had a large performance drop between the test and validation data set mitigation against overfitting could be researched.
\section{Conclusion} \label{sec:Conclusion}

The trained deep neural networks can detect clothing attributes from different rowing associations of live-action images. 
Using heatmaps it is shown that the clothing attributes present information that the network uses to classify the images.
As shown in \cref{subsec:Heatmap} the networks without attention already are able to focus on the correct parts of the images. Using attention did not yield a performance improvement. \newline

As evidenced by our results, the attention module isn't improving the test time results of ResNet-11 network in our classification task. Based on the results obtained, it can be said that attention might not necessarily improve performance in the case of fine-grained classification.  Attention suffered significantly more from overfitting compared to the other networks.

\section{Acknowledgements} \label{sec:Acknowledgements}

The authors would like to acknowledge the contributions of Randy Prozee gaining the high accuracy results of VGG networks and Martijn de Rooij for helping with the implementation of heatmaps.


{\small
\bibliographystyle{ieee}
\bibliography{references.bib,egbib.bib}
}

\newpage 
\section{Dataset Appendix}\label{FirstAppendix}

\begin{figure}[th!]
  \includegraphics[width=0.46\linewidth,keepaspectratio]{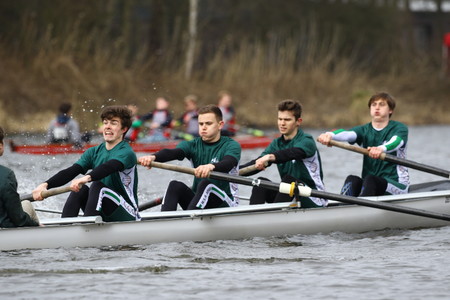}
  \includegraphics[width=0.46\linewidth,keepaspectratio]{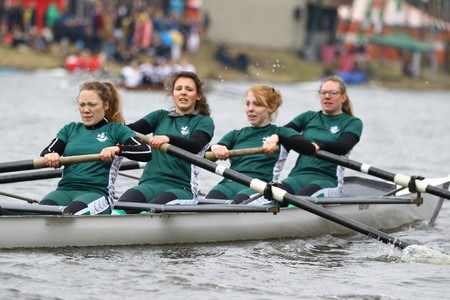} \\
  \includegraphics[width=0.46\linewidth,keepaspectratio]{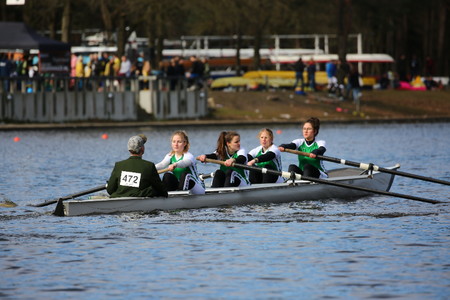}
  \includegraphics[width=0.46\linewidth,keepaspectratio]{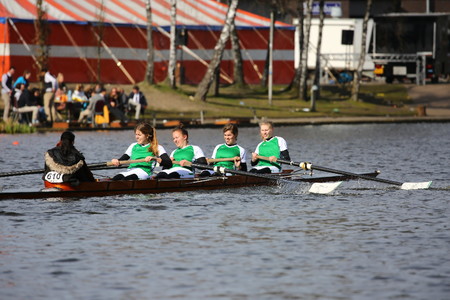}
  \caption{Argo}
  \label{fig:Argo Dataset}
\end{figure}

\begin{figure}[th!]
  \includegraphics[width=0.46\linewidth,keepaspectratio]{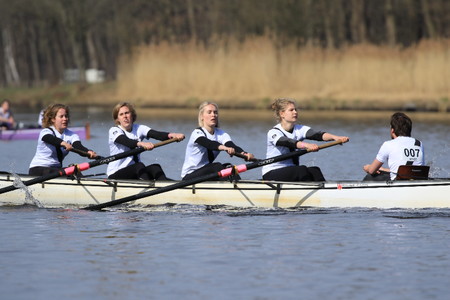}
  \includegraphics[width=0.46\linewidth,keepaspectratio]{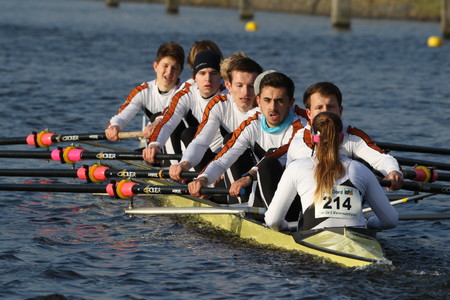} \\
  \includegraphics[width=0.46\linewidth,keepaspectratio]{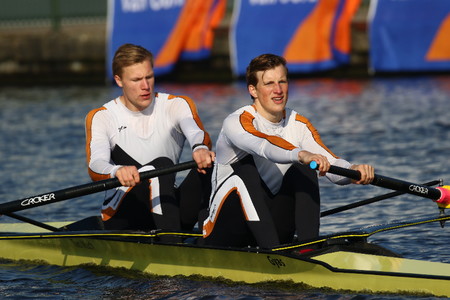}
  \includegraphics[width=0.46\linewidth,keepaspectratio]{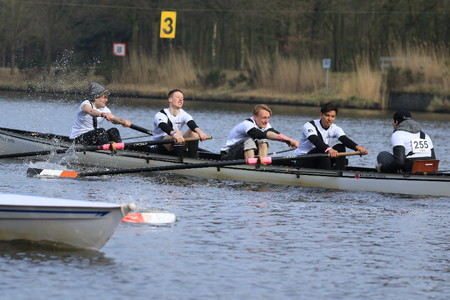}
  \caption{Gyas}
  \label{fig:Gyas Dataset}
\end{figure}

\begin{figure}[th!]
  \includegraphics[width=0.46\linewidth,keepaspectratio]{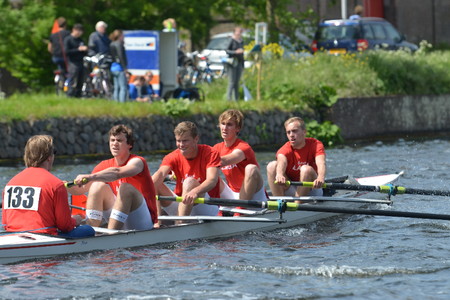}
  \includegraphics[width=0.46\linewidth,keepaspectratio]{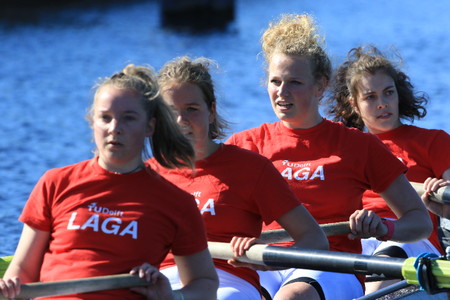} \\
  \includegraphics[width=0.46\linewidth,keepaspectratio]{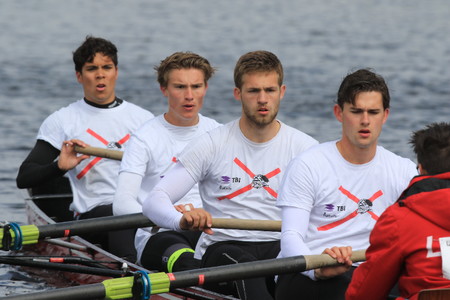}
  \includegraphics[width=0.46\linewidth,keepaspectratio]{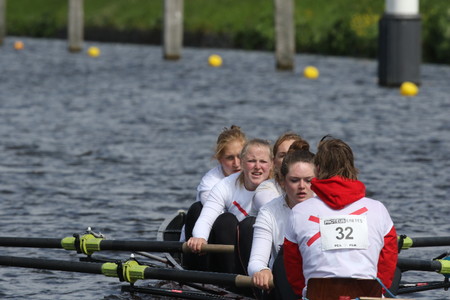}
  \caption{Laga}
  \label{fig:Laga Dataset}
\end{figure}

\begin{figure}[th!]
  \includegraphics[width=0.46\linewidth,keepaspectratio]{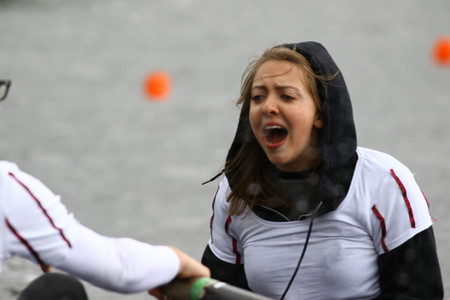}
  \includegraphics[width=0.46\linewidth,keepaspectratio]{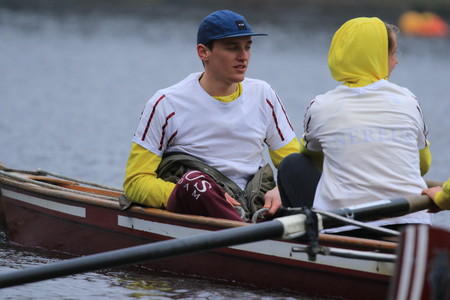} \\
  \includegraphics[width=0.46\linewidth,keepaspectratio]{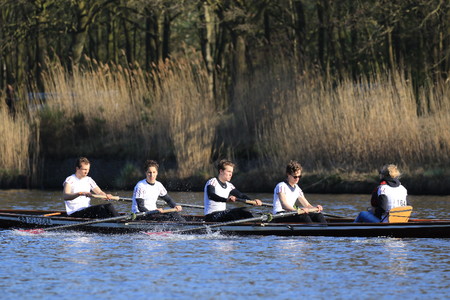}
  \includegraphics[width=0.46\linewidth,keepaspectratio]{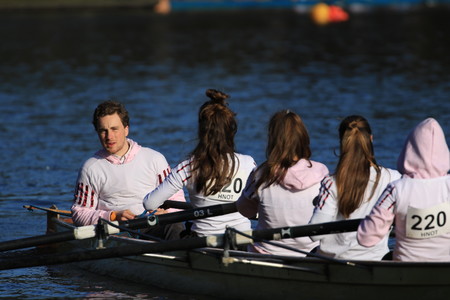}
  \caption{Nereus}
  \label{fig:Nereus Dataset}
\end{figure}

\begin{figure}[th!]
  \includegraphics[width=0.46\linewidth,keepaspectratio]{figures/Dataset/Njord/Njord1.jpg}
  \includegraphics[width=0.46\linewidth,keepaspectratio]{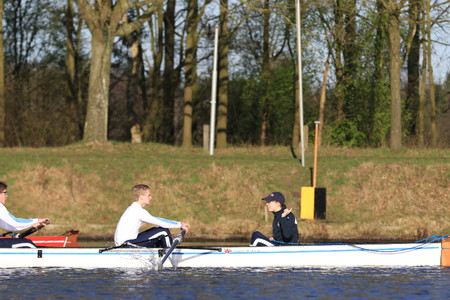} \\
  \includegraphics[width=0.46\linewidth,keepaspectratio]{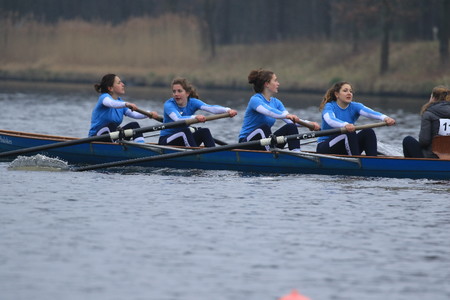}
  \includegraphics[width=0.46\linewidth,keepaspectratio]{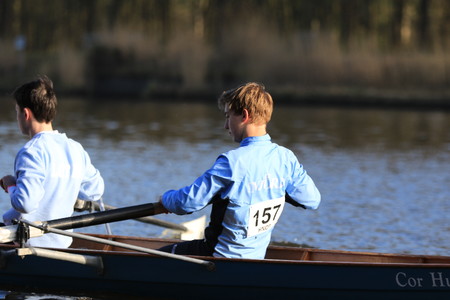}
  \caption{Njord}
  \label{fig:Njord Dataset}
\end{figure}

\begin{figure}[th!]
  \includegraphics[width=0.46\linewidth,keepaspectratio]{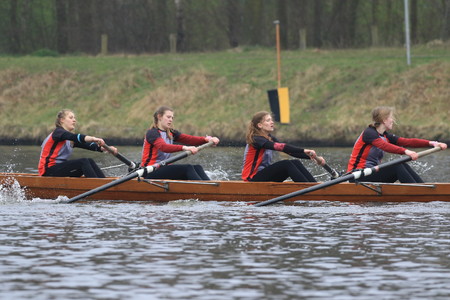}
  \includegraphics[width=0.46\linewidth,keepaspectratio]{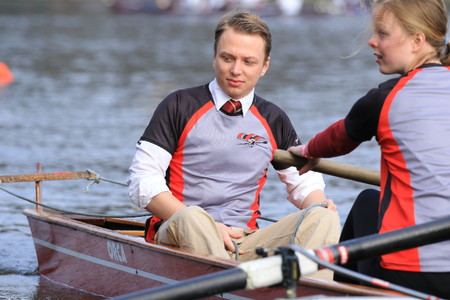} \\
  \includegraphics[width=0.46\linewidth,keepaspectratio]{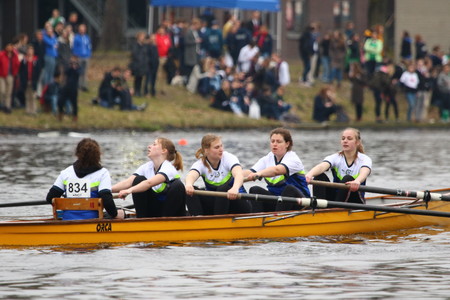}
  \includegraphics[width=0.46\linewidth,keepaspectratio]{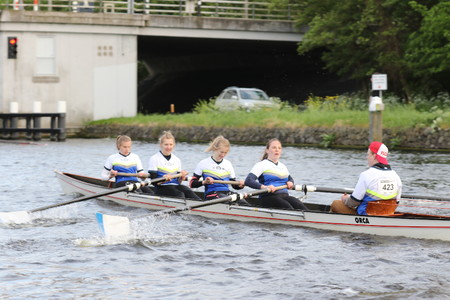}
  \caption{Orca}
  \label{fig:Orca Dataset}
\end{figure}

\begin{figure}[th!]
  \includegraphics[width=0.46\linewidth,keepaspectratio]{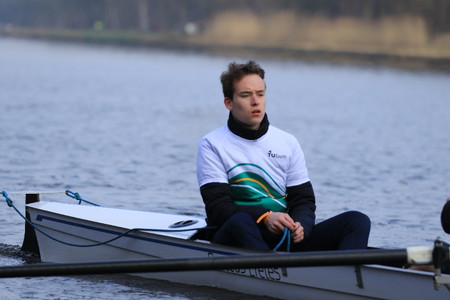}
  \includegraphics[width=0.46\linewidth,keepaspectratio]{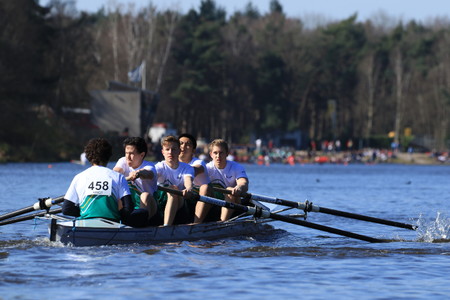} \\
  \includegraphics[width=0.46\linewidth,keepaspectratio]{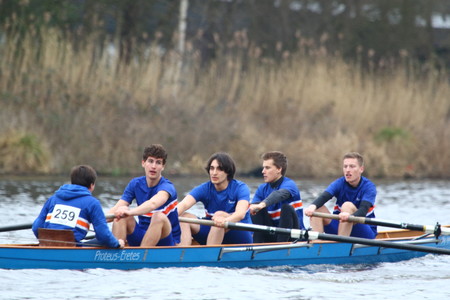}
  \includegraphics[width=0.46\linewidth,keepaspectratio]{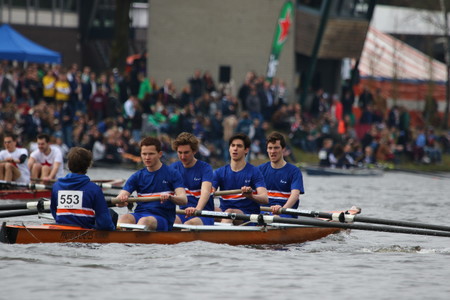}
  \caption{\pe}
  \label{fig:Proteus Dataset}
\end{figure}

\begin{figure}[th!]
  \includegraphics[width=0.46\linewidth,keepaspectratio]{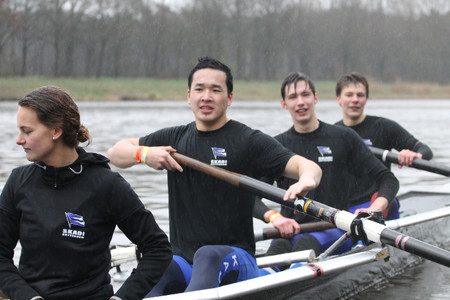}
  \includegraphics[width=0.46\linewidth,keepaspectratio]{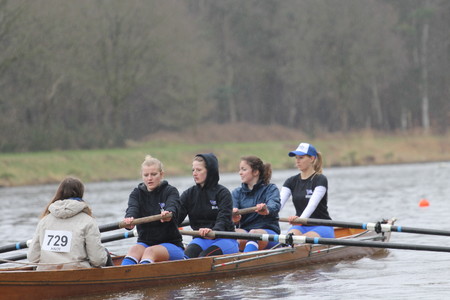} \\
  \includegraphics[width=0.46\linewidth,keepaspectratio]{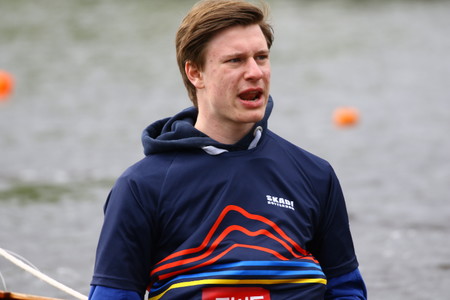}
  \includegraphics[width=0.46\linewidth,keepaspectratio]{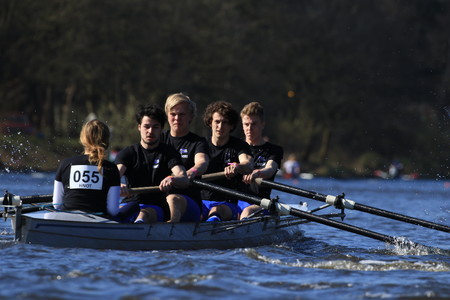}
  \caption{Skadi}
  \label{fig:Skadi Dataset}
\end{figure}

\begin{figure}[th!]
  \includegraphics[width=0.46\linewidth,keepaspectratio]{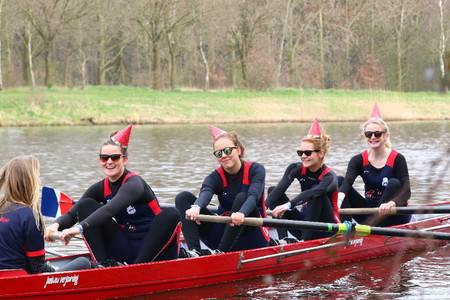}
  \includegraphics[width=0.46\linewidth,keepaspectratio]{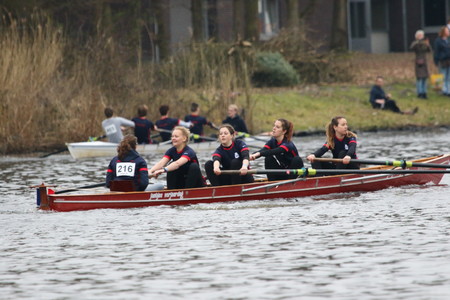} \\
  \includegraphics[width=0.46\linewidth,keepaspectratio]{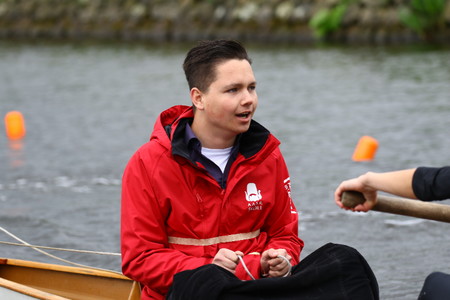}
  \includegraphics[width=0.46\linewidth,keepaspectratio]{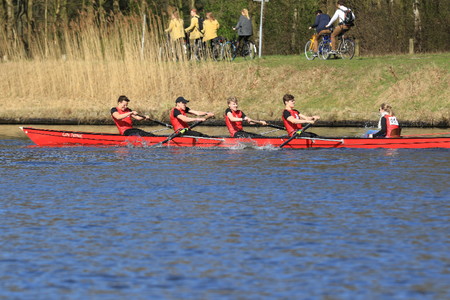}
  \caption{Skoll}
  \label{fig:Skoll Dataset}
\end{figure}

\begin{figure}[th!]
  \includegraphics[width=0.46\linewidth,keepaspectratio]{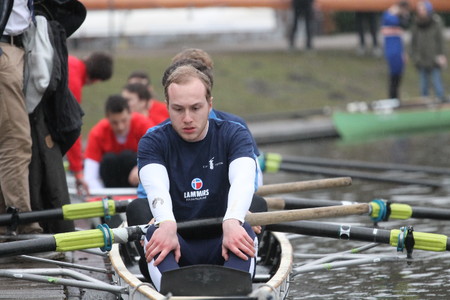}
  \includegraphics[width=0.46\linewidth,keepaspectratio]{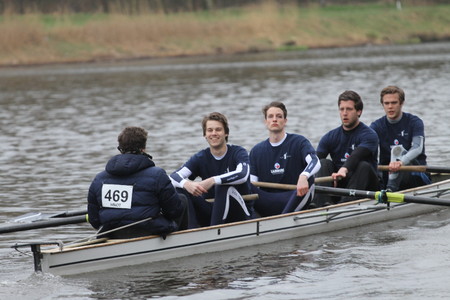} \\
  \includegraphics[width=0.46\linewidth,keepaspectratio]{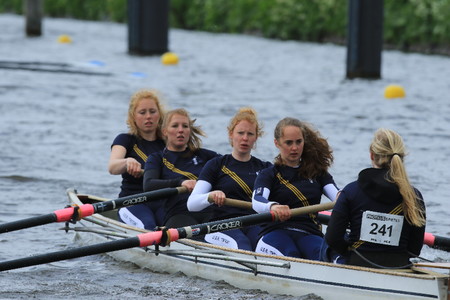}
  \includegraphics[width=0.46\linewidth,keepaspectratio]{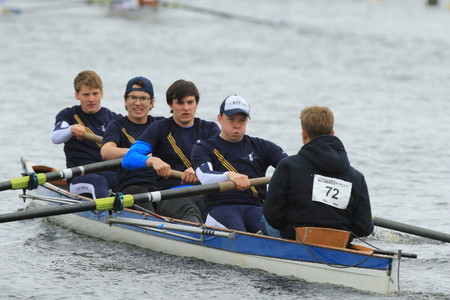}
  \caption{Triton}
  \label{fig:Triton Dataset}
\end{figure}

\end{document}